\documentclass[11pt]{article}
\PassOptionsToPackage{table,xcdraw,usenames,dvipsnames}{xcolor}
\usepackage[final]{acl}
\usepackage{times}
\usepackage{latexsym}
\usepackage{xurl}
\usepackage{amsmath,amsfonts,bm}

\def\eqref#1{equation~\ref{#1}}
\def\1{\bm{1}}

\DeclareMathAlphabet{\mathsfit}{\encodingdefault}{\sfdefault}{m}{sl}
\SetMathAlphabet{\mathsfit}{bold}{\encodingdefault}{\sfdefault}{bx}{n}

\usepackage[utf8]{inputenc} % allow utf-8 input
\usepackage[T1]{fontenc}    % use 8-bit T1 fonts
\usepackage{hyperref}       % hyperlinks
\usepackage{url}            % simple URL typesetting
\usepackage{booktabs}       % professional-quality tables
\usepackage{colortbl}       % colored tables
\usepackage{amsthm}         % theorem/proposition environments
\usepackage{amsfonts}       % blackboard math symbols
\usepackage{amsmath}
\usepackage{amssymb}
\usepackage{amsthm}         % theorem/proof environments
\usepackage{nicefrac}       % compact symbols for 1/2, etc.
\usepackage{microtype}      % microtypography
\usepackage{xcolor}
\usepackage{multirow}
\usepackage{array}
\usepackage{wrapfig}
\usepackage{tikz}
\usepackage[ruled,linesnumbered]{algorithm2e}
\usepackage{graphicx}
\usepackage{enumerate}
\usepackage{enumitem}
\usepackage[most]{tcolorbox}
\usepackage{xspace}
\usepackage[normalem]{ulem}
\usepackage{makecell}       % provides \Xhline
\usepackage{bbding}
\usepackage{siunitx}
\usepackage{caption}
\usepackage{subcaption}
\usepackage{listings}
\usepackage{pifont}
\usepackage{fontawesome5}
\usepackage{cleveref}
\crefname{algocf}{Algorithm}{Algorithms}
\Crefname{algocf}{Algorithm}{Algorithms}

\newtheorem{definition}{Definition}

\usepackage{booktabs}
\usepackage{tabularx}
\usepackage{array}
\usepackage{enumitem}
\usepackage{tcolorbox}
\tcbuselibrary{breakable}

\tcbset{
  agentbox/.style={
    enhanced=false,
    breakable,
    colback=black!3,
    colframe=black!20,
    colbacktitle=black!8,
    coltitle=black,
    fonttitle=\small\bfseries,
    fontupper=\small,
    boxrule=0.4pt,
    arc=0.8mm,
    left=5pt,
    right=5pt,
    top=4pt,
    bottom=4pt,
    before skip=6pt,
    after skip=6pt
  }
}

\hypersetup{
    colorlinks=true,
    linkcolor=red,
    citecolor=cyan,
    filecolor=magenta,
    urlcolor=magenta,
}

\newcommand{\mycite}[1]{\hyperlink{cite.#1}{$\rhd$}}

\DeclareRobustCommand{\halfbulletLR}{%
  \raisebox{0.15ex}{%
    \begin{tikzpicture}[baseline=-0.5ex]
      \fill[black] (0,0) circle (0.5ex);
      \begin{scope}
        \clip (0,0) circle (0.5ex);
        \fill[white] (0,-0.5ex) rectangle (0.5ex,0.5ex);
      \end{scope}
      \draw[black] (0,0) circle (0.5ex);
    \end{tikzpicture}%
  }%
}

\newcommand{\ourmethod}{{\ttfamily{SIGMA}}\xspace}
\newcommand{\llmname}[1]{{\ttfamily{#1}}\xspace}
\newcommand{\blue}[1]{$_{\color{BlueGreen}\downarrow #1}$}
\newcommand{\red}[1]{$_{\color{RedOrange}\uparrow #1}$}
\definecolor{darksalmon}{rgb}{0.91, 0.59, 0.48}
\definecolor{mygrey}{gray}{0.4}
\definecolor{green(pigment)}{rgb}{0.0, 0.65, 0.31}

\SetKwComment{Comment}{/** }{ **/}
\SetKwInput{KwIn}{Input}
\SetKwInput{KwOut}{Output}
\SetAlFnt{\small}
\SetAlCapFnt{\small}
\SetAlCapNameFnt{\small}
\SetAlgoNlRelativeSize{-1}
\SetKwComment{tcp}{}{}

\lstdefinestyle{pythonstyle}{
    language=Python,
    basicstyle=\ttfamily\small,
    keywordstyle=\color{blue},
    commentstyle=\color{gray},
    stringstyle=\color{red!70!black},
    showstringspaces=false,
    numbers=none,
    numberstyle=\tiny,
    stepnumber=1,
    numbersep=5pt,
    frame=none,
    breaklines=true,
    tabsize=4
}
\title{Conflict-Aware Multi-Agent Reasoning via Signed Graph Modeling}

\author{
\textbf{Longgang He}$^{1}$ \quad
\textbf{Longzhu He}$^{2}$ \quad
\textbf{Shihao Peng}$^{1}$ \quad\\
\textbf{Daojing He}$^{1\dagger}$ \quad
\textbf{Chaozhuo Li}$^{2\dagger}$ \\
$^{1}$Harbin Institute of Technology (Shenzhen) \\
$^{2}$Beijing University of Posts and Telecommunications \\
\texttt{helonggang@stu.hit.edu.cn}\\
$^{\dagger}$Corresponding authors
}

\begin{document}

\maketitle

\begin{abstract} 
LLM-based multi-agent systems (MAS) seek to improve reasoning by combining diverse perspectives from multiple agents. However, aggregation may overlook supportive, contradictory, and uncertain relations among their responses. As a result, useful dissent can be suppressed, allowing an incorrect majority to dominate the final decision.
We introduce \ourmethod, a novel \textit{\underline{SI}gned \underline{G}raph-informed \underline{M}ulti-\underline{A}gent reasoning framework} that distinguishes support, contradiction, and uncertainty among responses from black-box LLM agents. \ourmethod constructs a signed evidence graph, reassesses disputed relations, and forms reasoning communities while preserving distinct evidence. It then conducts bounded cross-examination between community representatives and adjudicates their evidence without treating community size as a proxy for correctness.
Extensive experiments on six benchmark datasets, across multiple LLM backbones and diverse multi-agent configurations, demonstrate that \ourmethod consistently outperforms state-of-the-art baselines, achieving notable gains in both accuracy and conflict-resilient performance. 
\end{abstract}

\section{Introduction}\label{sec:intro}

Large Language Models (LLMs) have demonstrated strong capabilities in reasoning~\citep{1,2,3,4}, planning~\citep{5,6,7}, and decision-making~\citep{1,9,10}. These capabilities have supported LLM-driven agents in applications such as code generation~\citep{11,12}, data analysis~\citep{13}, and embodied intelligence~\citep{14}. Building on single-agent systems, LLM-based multi-agent systems (MAS) coordinate multiple agents to introduce diverse perspectives and candidate solutions~\citep{r1,16,17}. By combining agents with different knowledge and expertise~\citep{18,19}, MAS extend individual reasoning into collaborative problem solving.

\begin{figure}[t]
    \centering
    \includegraphics[width=\linewidth]{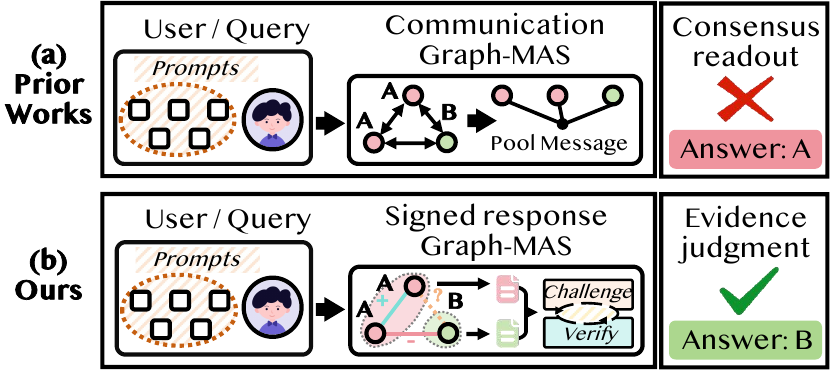}
    \caption{Comparison of prior work and our \ourmethod.
{(a)} Graph-based pooling may reinforce shared errors
and suppress minority answers.
{(b)} \ourmethod preserves competing hypotheses through
signed relations, tests assumptions through cross-examination,
and adjudicates answers based on evidence rather than consensus.}
    \label{fig:intro}
\end{figure}

Existing MAS methods adopt different interaction topologies, including \textit{chain-based}~\citep{r5,r6,Chatdev}, \textit{hierarchical}~\citep{37,38}, and \textit{graph-based}~\citep{39,41,43,GraphPlanner} structures. Chain-based methods organize agents into sequential workflows, whereas hierarchical methods coordinate outputs across multiple levels or branches. Graph-based methods represent agents as nodes and interactions as edges, supporting flexible information exchange and response aggregation~\citep{4,32,r7}. Their flexible connectivity accommodates complex dependencies and task-dependent interaction patterns. As illustrated in~\Cref{fig:intro}(a), a representative
workflow exchanges agent messages through a structured
communication graph and aggregates them into a
final collective decision.

However, existing graph-based multi-agent approaches still face two key limitations. First, pairwise links alone do not capture higher-order consistency among multiple responses. Consider three responses where A supports B, B supports C, but A contradicts C. Grouping them through the two support links would conceal the A–C conflict and could merge incompatible hypotheses into an apparent consensus. This motivates checking whether local relations compose into coherent communities before aggregation. Second, communication- or relevance-based graphs do not explicitly distinguish support, contradiction, and uncertainty among response claims. Agents sharing models, prompts, or reasoning biases may repeat the same error, while a dissenting response may contain correct evidence. Without explicit relation modeling, aggregation can mistake repeated opinions for independent evidence and suppress useful dissent.

To address these limitations, we propose \ourmethod, a \textit{\underline{SI}gned \underline{G}raph-informed \underline{M}ulti-\underline{A}gent Reasoning} framework. Drawing on signed graph modeling~\citep{signd,r12,r13,r14}, \ourmethod represents response instances as nodes and explicitly distinguishes support, contradiction, and uncertainty. As shown in~\Cref{fig:intro}(b), it uses higher-order structural conflicts to identify relations that warrant reassessment. These relations organize competing hypotheses rather than label agents as reliable or answers as correct. The framework preserves hypotheses and evidence through reasoning communities, then guides bounded cross-examination between representatives before evidence-based adjudication.

Implementing this framework involves three modeling challenges.
\ding{172} \textit{Relation inference and higher-order consistency checking.}
For black-box LLM agents, relations must be inferred from observable answers and reasoning text. The system must also identify potential conflicts among local judgments without inferring additional relations through sign propagation.
\ding{173} \textit{Joint compression and hypothesis preservation.}
Community formation and representative selection must reduce redundancy while retaining potentially valuable minority hypotheses and evidence not covered by the representatives.
\ding{174} \textit{Budgeted cross-examination and adjudication.}
The system must select disputes worth examining under a limited interaction budget. Its adjudicator must evaluate concrete evidence without treating community size or repeated agreement as a proxy for correctness.

To address these challenges, \ourmethod selects relevant agents with complementary capabilities to independently generate answers and rationales. For non-unanimous responses, it builds a signed evidence graph and uses uncertainty, community boundaries, and higher-order conflicts to guide budgeted reassessment of disputed pairs rather than sign propagation. It then forms reasoning communities, selects representatives, and preserves uncovered evidence in community packets. Representatives conduct bounded challenge--reply exchanges over selected disputes. Finally, an evidence-based adjudicator integrates their responses, community evidence, debate records, and unresolved relations without treating community size as evidence of correctness. The signed graph guides hypothesis organization, dispute localization, and interaction scheduling throughout the process.

\textbf{Contributions.} We summarize our main contributions as follows:
\ding{172} We define disagreement-aware multi-agent aggregation, specifying the roles of response-relation modeling, higher-order consistency checking, and competing-hypothesis preservation.
\ding{173} We propose \ourmethod, which uses a signed evidence graph to guide relation reassessment, reasoning communities, evidence preservation, and bounded cross-examination before adjudication.
\ding{174} We evaluate \ourmethod under shared initial responses and analyze accuracy, inference cost, and matched reasoning trajectories, including correct-minority recovery and harmful revisions.

\textbf{Organization.} \Cref{sec:preliminaries} introduces the preliminaries and signed graph formulation. \Cref{sec:methodology} presents \ourmethod, including agent selection, graph construction and auditing, community interaction, and evidence-based adjudication. \Cref{sec:exp} reports the experimental results and analyses. \Cref{secrw} reviews related work, and \Cref{secc} concludes the paper. 
Finally, the Limitations section discusses the evaluation scope and potential deployment risks.

\section{Preliminaries}
\label{sec:preliminaries}

This section formalizes LLM-based multi-agent reasoning and defines
a signed evidence graph over observable response instances.

\paragraph{LLM-based Multi-Agent System Formalization.}\label{gai1}
An LLM-based multi-agent system can be represented by a directed
communication graph $\mathcal{G}=(\mathcal{V},\mathcal{E})$,
where $\mathcal{V}$ contains agents and $\mathcal{E}$ contains
communication links~\citep{1,2,r1}.
Each agent is abstracted as:
\begin{equation}
v_i=(\mathcal{M}_i,\mathcal{R}_i,\mathcal{P}_i),
\end{equation}
where $\mathcal{M}_i$ is its underlying model,
$\mathcal{R}_i$ specifies its role, and $\mathcal{P}_i$
defines its prompting strategy.

Given a query $Q$, the system selects up to $k$ response slots.
A model may occupy multiple slots, so the number of responses
need not equal the number of distinct model checkpoints.
Each slot independently produces an initial response
\(x_i=(a_i,\rho_i,c_i)\),
where $a_i$ is the parsed answer, $\rho_i$ is the observable
response text, and $c_i$ is optional self-reported confidence.
Let $X=\{x_1,\ldots,x_n\}$, where $n\leq k$.
Neither self-reported confidence nor repeated agreement establishes correctness.

We define disagreement-aware aggregation as producing a final
prediction from $Q$ and $X$ while explicitly organizing response
relations, preserving competing evidence, and allocating bounded
interaction. Relations attach to responses rather than permanent agent identities. Initial responses remain available when interactions revise answers.

\paragraph{Signed Evidence Graph.}
Drawing on signed graph modeling~\citep{signd,r13,r14},
\ourmethod constructs \(G_Q=(X,E^{+},E^{-},E^{?})\),
where $E^{+}$ contains supportive response pairs,
$E^{-}$ contains contradictory pairs, and $E^{?}$
contains pairs whose relations remain uncertain.
For non-unanimous responses, the implemented pipeline
assesses every unordered pair, so these sets partition
the response pairs. Uncertainty denotes an inconclusive
assessment, not an unexamined pair.
The graph thus distinguishes support, conflict, and
uncertainty rather than reducing all interactions
to a single relevance score.

For each pair $(i,j)$, the assessor returns
\(e_{ij}=(\ell_{ij},\gamma_{ij},\xi_{ij})\),
where $\ell_{ij}\in\{+,-,?\}$ is a categorical relation
label, $\gamma_{ij}\in[0,1]$ is confidence in that label,
and $\xi_{ij}$ is a response-grounded rationale.
These records are relation judgments, not probabilities
of answer correctness.
Support requires compatible claim-level reasoning;
contradiction requires an identifiable incompatible claim.
Mixed comparisons or insufficient evidence remain uncertain.
This conservative labeling preserves ambiguous evidence
for subsequent examination.
The symmetric signed matrix is:
\begin{equation}
A_{ij}=
\begin{cases}
+\gamma_{ij}, & \ell_{ij}=+,\\
-\gamma_{ij}, & \ell_{ij}=-,\\
0, & \ell_{ij}=?,
\end{cases}
\qquad A_{ii}=0.
\label{eq:signed-matrix}
\end{equation}
A positive edge does not prove either response correct,
and a negative edge does not identify an unreliable agent.
Uncertain pairs remain explicitly recorded despite their
zero matrix entries.

\begin{tcolorbox}[
  colback=black!5,
  colframe=black!5,
  boxrule=0pt,
  arc=0.5mm,
  left=2pt, right=2pt, top=2pt, bottom=2pt
]
\begin{definition}[Signed Triad Balance~\citep{heider1946attitudes,signd}]
\label{def:balance-theory}
Consider three responses whose pairwise relations are all
decisive, with signs $s_{ij},s_{jt},s_{it}\in\{-1,+1\}$.
The triad is balanced if
\(
s_{ij}s_{jt}s_{it}=+1\),
and imbalanced otherwise.
Triads containing uncertain relations are not assigned a
balance status.
\end{definition}
\end{tcolorbox}

In \ourmethod, higher-order checking refers to this triadic
diagnostic over directly assessed pairwise relations.
An imbalanced triad identifies a structural tension worth
reassessing, not an erroneous edge or an incorrect answer.
In particular, local support between A and B, and between
B and C, does not determine the A--C relation.
The system therefore reassesses selected pairs directly
rather than propagating edge signs.

To organize competing hypotheses, let
$\mathcal{C}=\{C_1,\ldots,C_M\}$ partition the responses,
and let $z_i$ denote the community containing $x_i$.
We use the weighted disagreement objective:
\begin{equation}
\begin{aligned}
\mathcal{L}_{\mathrm{cc}}(\mathcal{C})
&=\sum\nolimits_{(i,j)\in E^{+}}
  \gamma_{ij}\mathbb{I}[z_i\neq z_j]\\
&\quad+\sum\nolimits_{(i,j)\in E^{-}}
  \gamma_{ij}\mathbb{I}[z_i=z_j].
\end{aligned}
\label{eq:cc}
\end{equation}
The objective penalizes separating supportive responses
and merging contradictory responses.
Uncertain pairs contribute no penalty.
Singleton communities are allowed, and community size is
not a correctness score.
Section~\ref{sec:methodology} describes reassessment,
evidence preservation, and interaction.

\section{Methodology}
\label{sec:methodology}

\begin{figure*}[t]
    \centering
    \includegraphics[width=\textwidth]{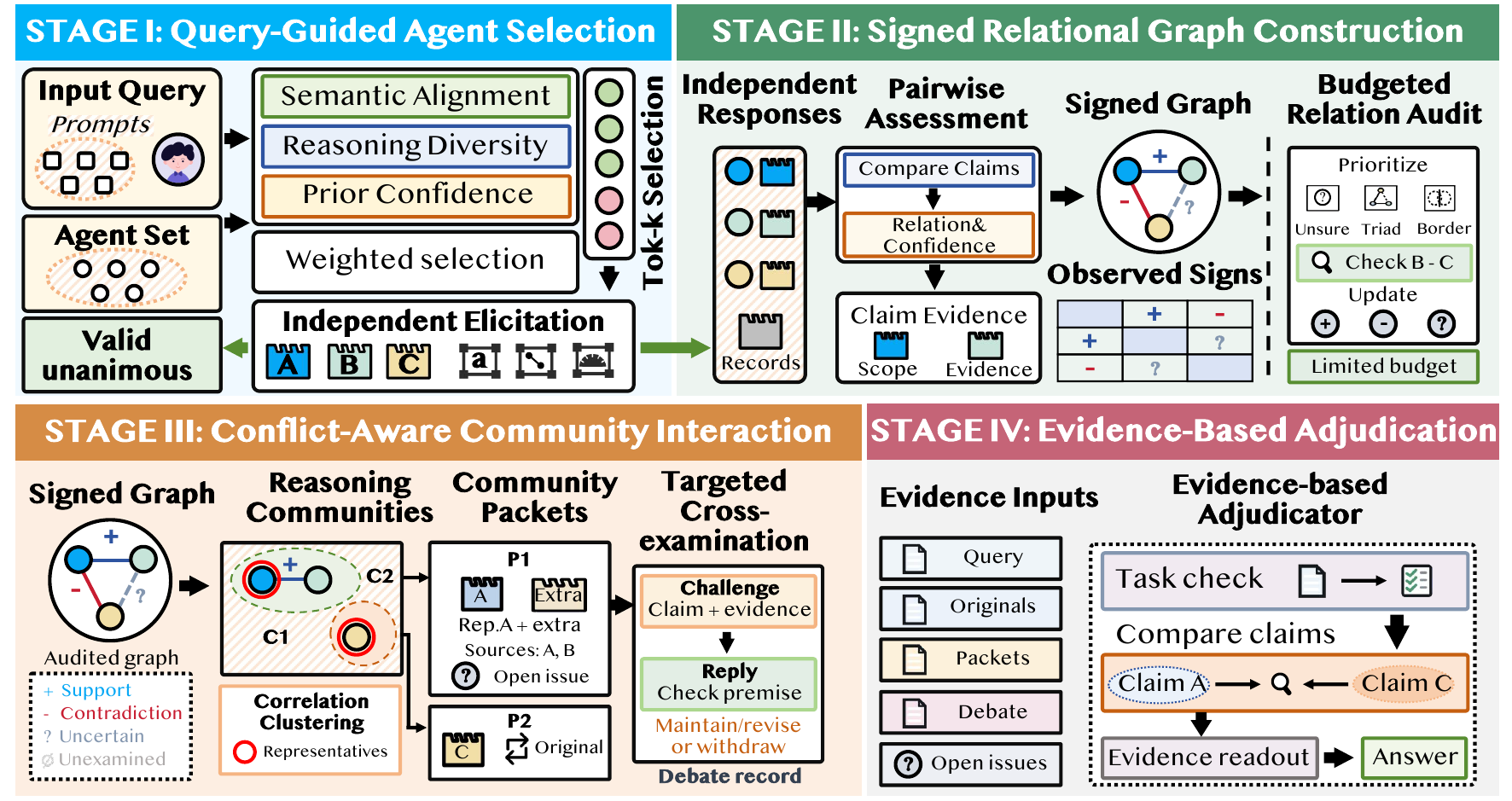}
    \caption{Overview of \ourmethod.
    (I) Query-guided selection collects independent responses.
    (II) A signed response graph captures support, contradiction,
    and uncertainty; conservative auditing independently reassesses
    selected relations.
    (III) Communities preserve original representatives
    and source-linked excerpts, followed by targeted cross-examination.
    (IV) An adjudicator compares retained evidence, debate
    records, and unresolved relations.}
    \label{fig:me}
\end{figure*}

As illustrated in~\Cref{fig:me}, \ourmethod organizes
multi-agent reasoning through four stages: Query-Guided Agent Selection ($\triangleright$ \Cref{sec:agent-selection}), Signed Relational Graph Construction ($\triangleright$ \Cref{sec:signed-graph}), Conflict-Aware Community Interaction ($\triangleright$ \Cref{sec:signed-propagation}), and Evidence-Based Adjudication ($\triangleright$ \Cref{sec:phase4}).

\subsection{Query-Guided Agent Selection}
\label{sec:agent-selection}

We adopt the model-card-based selector used by
GoA~\citep{41}.
Given a query and candidate model descriptions,
the selector prioritizes domain match, task specialization,
and the inclusion of a generalist where applicable.
It returns up to $k$ response slots:
\((v_{\sigma(1)},\ldots,v_{\sigma(n)})
=\mathsf{Select}(Q,\mathcal{D},k),
 n\leq k\),
where $\mathcal{D}$ contains model cards.
Repeated model selections are permitted.
Selection is performed before generating the current
responses and does not use their answer confidence.

Each selected slot independently generates $x_i=(a_i,\rho_i,c_i)$ without observing other responses.
The resulting cache is reused in matched aggregation comparisons, isolating post-response coordination from initial-response generation.
Low-confidence replacement is disabled in the reported local evaluation.
Before graph construction, \ourmethod checks whether all parsed answers are valid and identical:
\begin{equation}
\mathsf{Unanimous}(X)
=\mathbb{I}[a_1=\cdots=a_n\in\mathcal{A}].
\label{eq:unanimous-stop}
\end{equation}

where $\mathcal{A}$ denotes the set of valid answers.
If this condition holds, the system returns the first original
response without relation assessment, cross-examination,
or adjudication. This shortcut avoids inference when no
answer-level disagreement is present. It does not guarantee
correctness.

\subsection{Signed Relational Graph Construction}
\label{sec:signed-graph}

For non-unanimous responses, a relation assessor examines
all $\binom{n}{2}$ unordered pairs.
Given $Q$, $\rho_i$, and $\rho_j$, it returns the
categorical label, confidence, and rationale defined in
Section~\ref{sec:preliminaries}.
The prompt requests a decisive agreeing or conflicting
claim, or an explanation of missing evidence.
Matching final answers alone does not establish support.

Confidence values are clipped to $[0,1]$.
Malformed assessments default to uncertainty.
A claimed supportive or contradictory relation is also
converted to uncertainty when its rationale is empty
or its confidence is below $\tau=0.5$.
Accepted decisive relations use confidence directly as
their edge magnitude, as in Eq.~\ref{eq:signed-matrix}.

\noindent $\blacksquare$\, \textbf{Higher-order structural checking.}
The system examines complete decisive triads.
For each edge:
\begin{equation}
t_{ij}=
\mathbb{I}\!\left[
\begin{array}{l}
\exists h\notin\{i,j\}:
s_{ij},s_{ih},s_{jh}\in\{-1,+1\},\\
s_{ij}s_{ih}s_{jh}=-1
\end{array}
\right].
\label{eq:triad-tension}
\end{equation}
This indicator marks participation in an imbalanced
triad and supplies a reassessment cue.
It neither identifies which judgment is wrong nor
forces the graph to become balanced.

\noindent $\blacksquare$\, \textbf{Budgeted relation reassessment.}
The implemented priority score is
$\pi_{ij}=u_{ij}+t_{ij}+b_{ij}+2d_{ij}$,
where $u_{ij}=1$ for $\ell_{ij}=?$ and
$u_{ij}=1-\gamma_{ij}$ for $\ell_{ij}\in\{+,-\}$,
and $d_{ij}=\mathbb{I}[a_i\neq a_j]$.
Here $b_{ij}$ indicates whether removing the pair's
relation changes a provisional partition.
For this boundary diagnostic, communities are additionally
constrained not to merge distinct valid parsed answers.
The diagnostic partition is separate from the final
partitioning step described below.

The highest-priority pairs are independently reassessed,
up to a budget $B_{\mathrm{rel}}$.
The reassessment prompt considers support, contradiction,
and uncertainty using the original response texts,
without inferring a label from other pairs.
The default local budget is $B_{\mathrm{rel}}=1$.
Let $(\ell_{ij},\gamma_{ij})$ and
$(\ell'_{ij},\gamma'_{ij})$ be the initial and reassessed
judgments. They are combined conservatively:
\begin{equation}
\widetilde{\ell}_{ij}=
\begin{cases}
\ell_{ij},
& \ell_{ij}=\ell'_{ij}\in\{+,-\},\\
?,
& \text{otherwise},
\end{cases}
\label{eq:conservative-audit}
\end{equation}
with
$\widetilde{\gamma}_{ij}
=\min(\gamma_{ij},\gamma'_{ij})$.
Thus, conflicting judgments remain uncertain rather
than being resolved by selecting the latest label.

After reassessment, an answer-consistency safeguard
converts support to uncertainty when both parsed answers
are valid but different.
This rule preserves the disagreement for subsequent
processing; it does not automatically create a
contradiction edge.
The updated labels and confidences define the graph
used for final community formation.

\subsection{Conflict-Aware Community Interaction}
\label{sec:signed-propagation}

\noindent $\blacksquare$\, \textbf{Community formation.}
We minimize Eq.~\ref{eq:cc} by enumerating set partitions
for the small response pool.
The implementation supports up to eight responses,
and the reported local configuration uses at most three.
Among equal-loss partitions, it prefers more communities,
followed by a deterministic ordering.
This tie-break avoids merging responses solely because
their uncertain relations impose no penalty.
The communities organize hypotheses rather than assign correctness labels.

\noindent $\blacksquare$\, \textbf{Representative selection and evidence preservation.}
For each community $C_m$, we select a graph medoid as its representative:
\begin{equation}
r_m=
\operatorname*{arg\,max}\nolimits_{i\in C_m}
\sum\nolimits_{\substack{j\in C_m\\j\neq i}} A_{ij}.
\label{eq:representative}
\end{equation}
Ties are resolved by response index.
The score measures within-community signed connectivity,
not calibrated reliability or answer correctness.
A community packet is
\((P_m=(r_m,U_m)\),
where $U_m$ contains source-linked excerpts from
non-representative members.
For non-singleton communities, an extraction prompt
requests up to four distinct items not covered by the
representative.
Each item must identify its source and reproduce an
exact contiguous quote from that response.

The implementation checks source membership and literal
quotation containment, retaining at most 500 characters
per item.
If extraction produces no accepted item, it retains
distinct source excerpts from non-representative members
as a fallback.
These checks establish textual provenance, not
correctness or complete semantic coverage.
Singleton communities retain their original response
without an additional extraction call.

\noindent $\blacksquare$\, \textbf{Cross-community scheduling.}
For communities $C_m$ and $C_n$, let
$\mathcal{A}_m=\{a_i:i\in C_m\}$.
Their interaction priority is:
\begin{equation}
\begin{aligned}
\omega_{mn}
&=\frac{1}{|C_m||C_n|}
\sum_{\substack{i\in C_m\\j\in C_n}}
\left(
\gamma_{ij}\mathbb{I}[\ell_{ij}=-]
+\mathbb{I}[\ell_{ij}=?]
\right)\\
&\quad+\mathbb{I}[\mathcal{A}_m\neq\mathcal{A}_n].
\end{aligned}
\label{eq:debate-priority}
\end{equation}
The normalized terms prioritize contradiction and
uncertainty without automatically favoring larger
communities.
The answer-set term prioritizes differing candidate
answers.

The system selects up to $B_{\mathrm{debate}}$
community pairs, with a default local budget of one.
Each selected pair conducts one challenge--reply exchange,
requiring two generation calls.
The smaller community initiates the challenge;
ties follow the deterministic pair order.

The challenger receives the query and both representatives'
original responses.
It identifies a concrete disputed assumption,
calculation, or claim and states what evidence would
change its own answer.
The respondent receives its original response and
the challenge, then explains whether it retains or
revises its answer.
Community evidence packets are retained for final
adjudication rather than supplied to these debate prompts.
The record
\(D_{mn}=(r_m,r_n,\chi_{mn},\psi_{mn})\)
stores the representative identities, challenge
$\chi_{mn}$, and reply $\psi_{mn}$.
The original responses remain unchanged in the cache.
The implementation does not reconstruct the relation
graph after the exchange.

\subsection{Evidence-Based Adjudication}
\label{sec:phase4}

The adjudicator receives one original representative
response per community, the retained evidence excerpts,
the cross-examination records, and the unresolved
uncertain pairs:
\begin{equation}
\widehat y=
J\!\left(
Q,\{\rho_{r_m},U_m\}_{m=1}^{M},
\{D_{mn}\},\mathcal{U}
\right),
\label{eq:adjudication}
\end{equation}
where
$\mathcal{U}=\{(i,j):\ell_{ij}=?\}$.
The prompt treats communities as alternative lines of
reasoning rather than votes.
It requests comparison against task constraints and
independent assessment of debate-induced revisions.
Community sizes, signed degrees, and vote counts are
not supplied as correctness weights.
All original responses are supplied in an ablation, not in the default adjudication input.

The default pipeline relies on model-generated
assessment and adjudication; it does not invoke
external execution or verification tools.
Consequently, preserving dissent and checking structural tensions can support evidence comparison without guaranteeing correct relations, valid reasoning, or a correct final answer.

\section{Experiment}\label{sec:exp}

We conduct a series of experiments to thoroughly evaluate the effectiveness of \ourmethod, first describing the experimental setup ($\triangleright$ \Cref{sec:exp-setup}) and then addressing the following five research questions:
\begin{itemize}[leftmargin=*, itemsep=-2pt]
    \item \textbf{\textit{RQ1:}} How does \ourmethod compare with different single- and multi-agent baselines? ($\triangleright$ \Cref{Main})
    \item \textbf{\textit{RQ2:}} What is the contribution of each component to \ourmethod's performance? ($\triangleright$ \Cref{ablation})
    \item \textbf{\textit{RQ3:}} Can \ourmethod recover minority answers while avoiding harmful revisions? ($\triangleright$ \Cref{robu})
    \item \textbf{\textit{RQ4:}} How sensitive is \ourmethod to hyperparameter settings and performance? ($\triangleright$ \Cref{hyperparam})
\item \textbf{\textit{RQ5:}} Can case studies demonstrate \ourmethod's modeling of trust and conflict? ($\triangleright$ \Cref{case})
\end{itemize}
\subsection{Experiment Setup}\label{sec:exp-setup}

\textbf{Datasets and Metrics.}
To evaluate the performance of \ourmethod, we conduct experiments on six benchmark datasets spanning two categories. The detailed statistics and characteristics of these datasets are summarized in~\Cref{tab:dataset_stats}. Specifically, the multi-domain category includes three general reasoning datasets: \texttt{MMLU}~\cite{mmlu}, \texttt{MMLU-Pro}~\cite{mmlu1}, and \texttt{GPQA}~\cite{gpqa}, covering knowledge across diverse subjects. The domain-specific category includes two mathematical datasets, \texttt{GSM8K}~\cite{cobbe} and \texttt{MultiArith}~\cite{he1}, and one coding dataset, \texttt{HumanEval}~\cite{he2}.

\textbf{Baselines.}
We select a diverse set of baselines for evaluation.
For single-agent approaches, we select {CoT} \cite{cot}, {ComplexCoT} \cite{ccot}, {Self-Consistency (SC)} \cite{r3}, and {PHP} \cite{php}.
For multi-agent approaches, we select {MoA} \cite{37}, {Self-MoA} \cite{38}, {Complete Graph}, {Random Graph}, {DyLAN} \cite{r10}, {AutoGen} \cite{r7}, {GPTSwarm} \cite{39}, {G-Designer} \cite{43}, and {GoA} \cite{41}.

\textbf{Implementation Details.}
We use the OpenAI API with
\llmname{gpt-5.4} (\llmname{GPT-5}) and
\llmname{gpt-5.4-mini} (\llmname{GPT-5 mini})
as backbones. Decoding temperatures are $0.7$
for sampling and $0$ for deterministic baselines.
The relational evaluator $f_{\rm eval}$
(\Cref{sec:signed-graph}) combines
\textsc{all-MiniLM-L6-v2} cosine similarity
($\mathcal{D}=384$), answer consistency,
and confidence.
Agent profiles are generated by \llmname{gpt-5.4}.
Our implementation uses text-based LLM interactions
and signed-weight aggregation without explicit
embedding-level neural message passing.

\begin{table*}[!t]
\centering
\label{tab:rq1_performance1}

\small
\setlength{\tabcolsep}{3pt}
\renewcommand{\arraystretch}{1.15}

\resizebox{\textwidth}{!}{
\begin{tabular}{lccc|ccccccc}
\toprule
\textbf{Method} &
\textbf{Mul.} &
\textbf{Rel.} &
\textbf{Conf.} &
\textbf{MMLU} &
\textbf{MMLU-Pro} &
\textbf{GPQA} &
\textbf{GSM8K} &
\textbf{MultiArith} &
\textbf{HumanEval} &
\textbf{Avg.} \\
\midrule

Vanilla
& \(\scalebox{1.3}{$\circ$}\)
& \(\scalebox{1.3}{$\circ$}\)
& \(\scalebox{1.3}{$\circ$}\)
& 89.58 & 84.17 & 45.96
& 92.80 & 97.42 & 93.75 & 83.95 \\

\rowcolor{gray!10}
CoT~\scalebox{0.75}{~(\mycite{cot} \textit{NeurIPS'22}$^\dagger$)}
& \(\scalebox{1.3}{$\circ$}\)
& \(\scalebox{1.3}{$\circ$}\)
& \(\scalebox{1.3}{$\circ$}\)
& 89.72\red{0.14}
& 87.58\red{3.41}
& 47.21\red{1.25}
& 94.47\red{1.67}
& 97.81\red{0.39}
& 95.25\red{1.50}
& 85.34 \\

ComplexCoT~\scalebox{0.75}{~(\mycite{scot} \textit{ICLR'23}$^\dagger$)}
& \(\scalebox{1.3}{$\circ$}\)
& \(\scalebox{1.3}{$\circ$}\)
& \(\scalebox{1.3}{$\circ$}\)
& 89.81\red{0.23}
& 88.76\red{4.59}
& 47.48\red{1.52}
& 94.74\red{1.94}
& 98.23\red{0.81}
& 95.10\red{1.35}
& 85.69 \\

\rowcolor{gray!10}
SC~\scalebox{0.75}{~(\mycite{m5} \textit{ICLR'23}$^\dagger$)}
& \(\scalebox{1.3}{$\circ$}\)
& \(\scalebox{1.3}{$\circ$}\)
& \(\scalebox{1.3}{$\circ$}\)
& 89.87\red{0.29}
& 91.43\red{7.26}
& 47.20\red{1.24}
& 94.76\red{1.96}
& 98.45\red{1.03}
& 95.37\red{1.62}
& 86.18 \\

PHP~\scalebox{0.75}{~(\mycite{php} \textit{ICML'24}$^\dagger$)}
& \(\scalebox{1.3}{$\bullet$}\)
& \(\scalebox{1.3}{$\circ$}\)
& \(\scalebox{1.3}{$\circ$}\)
& 90.22\red{0.64}
& 91.32\red{7.15}
& 48.32\red{2.36}
& 95.87\red{3.07}
& 98.50\red{1.08}
& 95.52\red{1.77}
& 86.62 \\

\midrule

\rowcolor{gray!10}
MoA~\scalebox{0.75}{~(\mycite{37} \textit{ICLR'25}$^\dagger$)}
& \(\scalebox{1.3}{$\bullet$}\)
& \(\scalebox{1.3}{$\circ$}\)
& \(\scalebox{1.3}{$\circ$}\)
& 89.93\red{0.35}
& 88.57\red{4.40}
& 52.54\red{6.58}
& 94.10\red{1.30}
& 98.22\red{0.80}
& 95.42\red{1.67}
& 86.46 \\

Self-MoA~\scalebox{0.75}{~(\mycite{38} \textit{arXiv'25}$^\dagger$)}
& \(\scalebox{1.3}{$\bullet$}\)
& \(\scalebox{1.3}{$\circ$}\)
& \(\scalebox{1.3}{$\circ$}\)
& 90.45\red{0.87}
& 88.69\red{4.52}
& 52.73\red{6.77}
& 94.34\red{1.54}
& 98.42\red{1.00}
& 95.48\red{1.73}
& 86.69 \\

\rowcolor{gray!10}
Complete Graph
& \(\scalebox{1.3}{$\bullet$}\)
& \(\scalebox{1.3}{$\bullet$}\)
& \(\scalebox{1.3}{$\circ$}\)
& 90.60\red{1.02}
& 94.73\red{10.56}
& 50.41\red{4.45}
& 94.21\red{1.41}
& 98.50\red{1.08}
& 95.82\red{2.07}
& 87.38 \\

Random Graph
& \(\scalebox{1.3}{$\bullet$}\)
& \(\scalebox{1.3}{$\bullet$}\)
& \(\scalebox{1.3}{$\circ$}\)
& 89.72\red{0.14}
& 93.42\red{9.25}
& 49.33\red{3.37}
& 94.15\red{1.35}
& 98.43\red{1.01}
& 95.10\red{1.35}
& 86.69 \\

\rowcolor{gray!10}
DyLAN~\scalebox{0.75}{~(\mycite{r10} \textit{arXiv'23}$^\dagger$)}
& \(\scalebox{1.3}{$\bullet$}\)
& \(\scalebox{1.3}{$\bullet$}\)
& \scalebox{1.1}{\halfbulletLR}
& 89.12\blue{0.46}
& 91.43\red{7.26}
& 49.72\red{3.76}
& 94.04\red{1.24}
& 97.54\red{0.10}
& 96.13\red{2.38}
& 86.33 \\

AutoGen~\scalebox{0.75}{~(\mycite{r7} \textit{COLM'24}$^\dagger$)}
& \(\scalebox{1.3}{$\bullet$}\)
& \(\scalebox{1.3}{$\bullet$}\)
& \scalebox{1.1}{\halfbulletLR}
& 89.31\blue{0.27}
& 93.29\red{9.12}
& 50.45\red{4.49}
& 95.13\red{2.33}
& 98.12\red{0.70}
& 96.01\red{2.26}
& 87.05 \\

\rowcolor{gray!10}
GPTSwarm~\scalebox{0.75}{~(\mycite{39} \textit{ICML'24}$^\dagger$)}
& \(\scalebox{1.3}{$\bullet$}\)
& \(\scalebox{1.3}{$\bullet$}\)
& \scalebox{1.1}{\halfbulletLR}
& 90.87\red{1.29}
& 94.87\red{10.70}
& 51.01\red{5.05}
& 95.31\red{2.51}
& 98.51\red{1.09}
& 96.45\red{2.70}
& 87.84 \\

G-Designer~\scalebox{0.75}{~(\mycite{43} \textit{ICML'25}$^\dagger$)}
& \(\scalebox{1.3}{$\bullet$}\)
& \(\scalebox{1.3}{$\bullet$}\)
& \scalebox{1.1}{\halfbulletLR}
& 91.44\red{1.86}
& \underline{95.31}\red{11.14}
& 52.72\red{6.76}
& \underline{96.10}\red{3.30}
& \underline{98.72}\red{1.30}
& \underline{96.88}\red{3.13}
& 88.53 \\

\rowcolor{gray!10}
GoA~\scalebox{0.75}{~(\mycite{1} \textit{ICLR'26})}
& \(\scalebox{1.3}{$\bullet$}\)
& \(\scalebox{1.3}{$\bullet$}\)
& \scalebox{1.1}{\halfbulletLR}
& \underline{91.73}\red{2.15}
& 95.10\red{10.93}
& \underline{53.16}\red{7.20}
& 95.25\red{2.45}
& 98.23\red{0.81}
& 96.32\red{2.57}
& 88.30 \\

\midrule

\rowcolor{gray!15}
\ourmethod
& \(\scalebox{1.3}{$\bullet$}\)
& \(\scalebox{1.3}{$\bullet$}\)
& \(\scalebox{1.3}{$\bullet$}\)
& \textbf{93.12}\red{3.54}
& \textbf{95.82}\red{11.65}
& \textbf{55.03}\red{9.07}
& \textbf{96.30}\red{3.50}
& \textbf{98.75}\red{1.33}
& \textbf{97.17}\red{3.42}
& \textbf{89.37} \\

\bottomrule
\end{tabular}
}
\caption{Performance comparison with single-agent and multi-agent baselines across multiple benchmarks. The best results are in \textbf{bold}, and the runner-ups are \underline{underlined}. ``Mul.'', ``Rel.'', and ``Conf.'' indicate multi-agent collaboration, inter-agent relation modeling, and conflict handling, respectively. 
\(\circ\), \halfbulletLR{}, and \(\bullet\) denote no, partial, and full support.
\(\dagger\) indicates papers with over 100 citations.
Colored subscripts denote absolute performance differences from Vanilla.}
\end{table*}

\subsection{Main Results (\textit{RQ1})}\label{Main}

We evaluate \ourmethod against single-agent and multi-agent
baselines across six benchmarks: three general reasoning datasets
(\texttt{MMLU}, \texttt{MMLU-Pro}, and \texttt{GPQA}) and three
domain-specific datasets (\texttt{GSM8K}, \texttt{MultiArith},
and \texttt{HumanEval}), as shown in \Cref{tab:rq1_performance1}.
Experiments use configurations based on
\llmname{gpt-5.4} (\llmname{GPT-5}) and
\llmname{gpt-5.4-mini} (\llmname{GPT-5 mini}).
\ourmethod achieves the best reported performance
on all six benchmarks, with an average score of \(89.37\)\%,
exceeding G-Designer and GoA by \(0.84\) and \(1.07\)
percentage points, respectively.
\ourmethod achieves \(93.12\)\% on \texttt{MMLU},
\(95.82\)\% on \texttt{MMLU-Pro}, \(55.03\)\% on \texttt{GPQA},
\(96.30\)\% on \texttt{GSM8K}, \(98.75\)\% on
\texttt{MultiArith}, and \(97.17\)\% on \texttt{HumanEval}.

Several graph-based baselines outperform single-agent approaches,
particularly on \texttt{MMLU-Pro} and \texttt{GPQA}, supporting
the value of structured collaboration.
Compared with the strongest baseline on each benchmark,
\ourmethod obtains its largest gains on \texttt{MMLU}
and \texttt{GPQA}, improving by \(1.39\) and \(1.87\)
percentage points, respectively, while improvements on
mathematical reasoning and code generation are smaller.
These results support the effectiveness of the overall framework,
which combines explicit support, contradiction, and uncertainty
modeling with conservative relation auditing, targeted
cross-examination, and evidence-based adjudication.

\begin{figure}[t]
    \centering
    \includegraphics[width=0.48\linewidth]{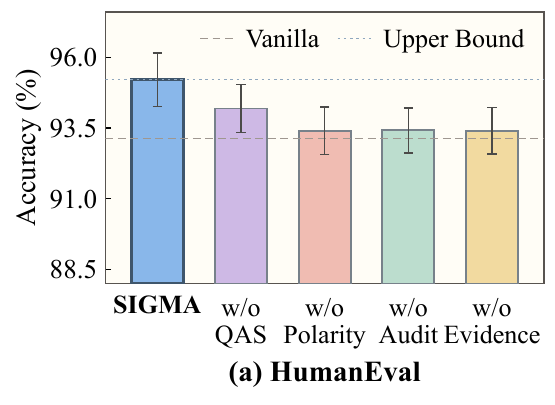}
    \hfill
    \includegraphics[width=0.48\linewidth]{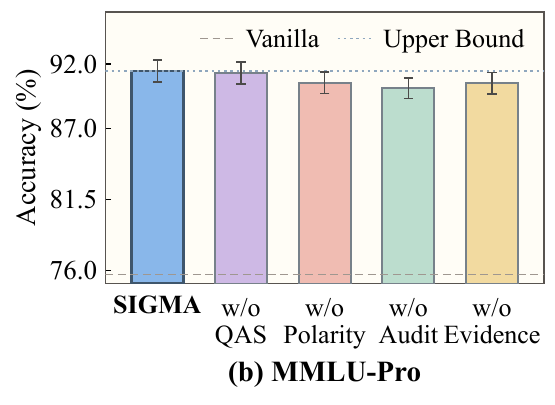}
    \caption{Ablation results of \ourmethod on
    HumanEval and MMLU-Pro.}
    \label{fig:ablation}
\end{figure}

\begin{figure}[h]
    \centering
    \includegraphics[width=0.48\linewidth]{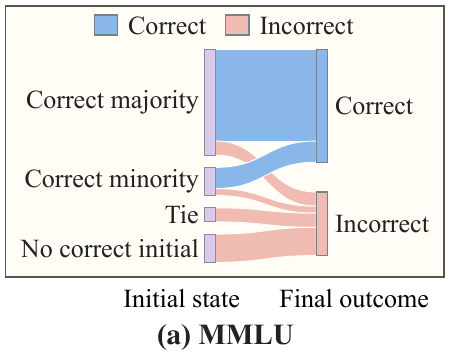}
    \hfill
    \includegraphics[width=0.48\linewidth]{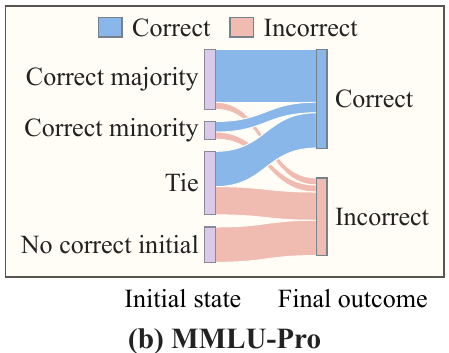}
    \caption{Initial answer composition and final
    outcomes of \ourmethod on MMLU and MMLU-Pro.}
    \label{fig:robustness}
\end{figure}

\subsection{Ablation Study (\textit{RQ2})}\label{ablation}

We conduct four ablation experiments, each using \llmname{DeepSeek-V3.2}
on \texttt{HumanEval} and \texttt{MMLU-Pro}:
\ding{172} \textit{w/o QAS} uses random selection
under the same response budget;
\ding{173} \textit{w/o Polarity} replaces audited
contradictions with uncertainty before community formation;
\ding{174} \textit{w/o Audit} disables relation reassessment
($B_{\mathrm{rel}}=0$);
\ding{175} \textit{w/o Evidence} removes supplementary
evidence while retaining community representatives.

As shown in \Cref{fig:ablation}, the full
\ourmethod performs best on both datasets.
Individual ablations generally reduce accuracy,
supporting the complementary contributions of
query-guided selection, contradiction modeling,
conservative auditing, evidence preservation.

\begin{figure}[h]
    \centering
    \includegraphics[width=0.8\linewidth]{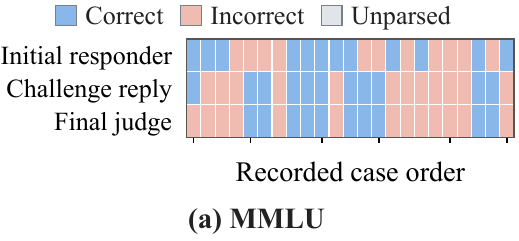}

    \vspace{0.5em}

    \includegraphics[width=0.8\linewidth]{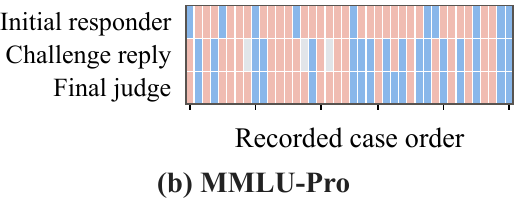}
    \caption{Answer trajectories of \ourmethod on
MMLU and MMLU-Pro. Columns trace cases from the
initial response through the challenge reply
to final judgment.}
    \label{fig:answer-trajectories1}
\end{figure}

\subsection{Disagreement and Answer Revision Analysis (RQ3)}
\label{robu}
We examine how \ourmethod handles initial answer disagreement and subsequent revisions on \texttt{MMLU} and \texttt{MMLU-Pro}.
Figure~\ref{fig:robustness} shows the final decisions under different initial answer compositions.
The results show that \ourmethod can recover correct minority answers and retain correct majority answers, enabling decisions that go beyond simple majority voting.
Figure~\ref{fig:answer-trajectories1} further traces answer changes across the initial response, challenge reply, and final adjudication.
Cross-examination can correct initially wrong answers, but it can also turn correct answers into incorrect ones.
These results demonstrate minority-answer recovery under natural disagreement and highlight the revision risks that must be controlled during interaction.

\begin{figure}[t]
    \centering
    \includegraphics[width=0.48\linewidth]{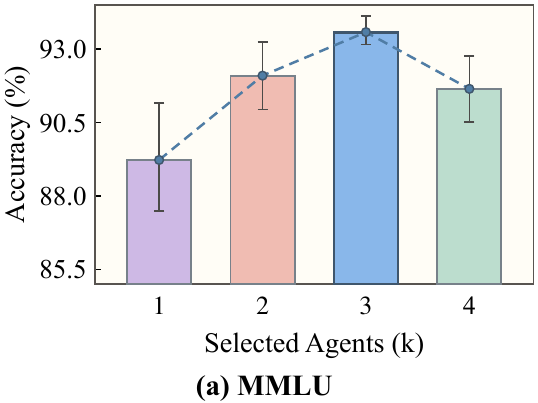}
    \hfill
    \includegraphics[width=0.48\linewidth]{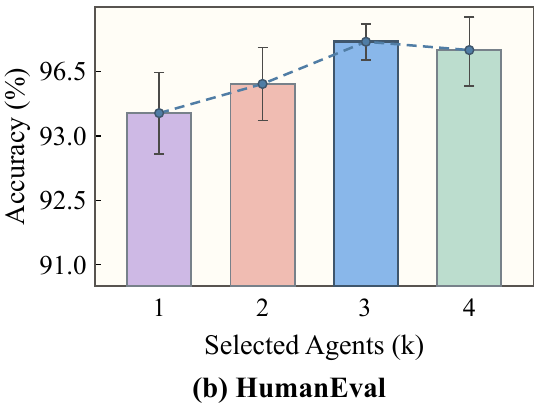}
    \caption{Sensitivity to the number of selected agents $k$ on (a) MMLU and (b) HumanEval.}
    \label{fig:k-sensitivity}
\end{figure}

\subsection{Hyperparameter Sensitivity (\textit{RQ4})}
\label{hyperparam}

Figure~\ref{fig:k-sensitivity} shows accuracy across
$k\in\{1,2,3,4\}$ on MMLU and HumanEval.
The plotted accuracy increases from $k=1$ to $3$
and declines at $k=4$ on both tasks.
Within the displayed range, $k=3$ achieves the
highest accuracy, suggesting that adding more
agents does not necessarily improve performance.
We use $k=3$ as the default configuration and fix
$B_{\mathrm{rel}}=B_{\mathrm{debate}}=1$,
allowing at most one relation audit and one
cross-community examination per query to bound
inference cost.

\subsection{Case Study (\textit{RQ5})}\label{case}

\begin{figure}[t] % r 表示右侧，宽度占页面宽度 25%
    \centering
    \includegraphics[width=0.9\linewidth]{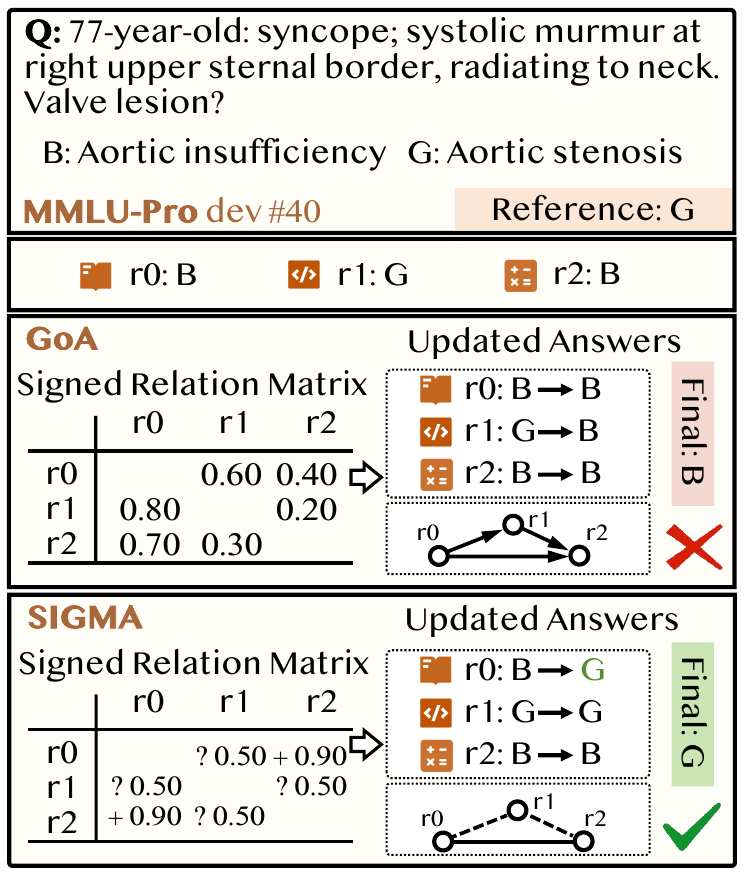} % 替换为你的图
    \caption{Detailed comparison on \texttt{MMLU-Pro} dev~\#40 (initial B/G/B; reference G). GoA-Max updates to B/B/B and returns B; \ourmethod updates to G/G/B and correctly returns G. GoA uses directed relevance scores, whereas \ourmethod uses symmetric signed relations ($?$: uncertainty; $+$: support; no contradiction edges).}
    \label{f33333} % label 紧跟在 caption 后面
\end{figure}

Figure~\ref{f33333} illustrates how \ourmethod preserves competing hypotheses and revises answers through cross-examination. In \texttt{MMLU-Pro} item~40, both methods start from B/G/B, with reference G. GoA updates the responses to B/B/B and incorrectly returns B, whereas \ourmethod updates them to G/G/B and correctly returns G, matching the reference answer. The supportive pair contains incorrect B responses, showing support does not imply correctness.

\section{Related Works}\label{secrw}

\textbf{LLM-based Multi-Agent Reasoning.}
LLM-based reasoning benefits from diverse responses
and collaborative interaction.
Representative approaches include multi-agent
debate~\cite{r1,r2} and single-model response
aggregation through Self-Consistency~\cite{r3}.
Subsequent frameworks introduce structured workflows
and communication, including MetaGPT~\cite{r5},
AutoGen~\cite{r7}, and hierarchical
organizations~\cite{r9}.
These approaches improve coordination, but
communication structure alone does not explicitly
distinguish support, contradiction, and uncertainty
among responses.

\textbf{Graphs for Multi-Agent Systems.}
Graphs support communication and coordination in multi-agent
reinforcement learning~\cite{marl1,marl2}.
For LLM-based systems~\cite{1,3}, DyLAN~\cite{r10},
GPTSwarm~\cite{39}, MacNet~\cite{r11},
G-Designer~\cite{43}, and MasRouter~\cite{44}
develop graph-based interaction or routing mechanisms.
Beyond organizing communication, response-level
relations distinguish agreement from
conflict and insufficient evidence.
Our approach models these relations
and combines conservative auditing, community
formation, and targeted debate for final adjudication.

\section{Conclusion}\label{secc}

In this paper, we investigate LLM-based MAS and identify the need to distinguish \textit{support}, \textit{contradiction}, and \textit{uncertainty} among agent responses, especially when their reasoning diverges.
To address this gap, we propose \ourmethod, a \textit{signed graph-informed multi-agent reasoning framework} that combines conservative relation auditing, community-based evidence preservation, and targeted debate to support final adjudication.
Experiments across reasoning benchmarks show improved accuracy and effective conflict handling.

\section{Limitations}

Our evaluation focuses on reasoning tasks with clear, verifiable answers, providing a controlled setting for examining agent relations, answer revisions, and final decisions. These tasks allow detailed analysis of how \ourmethod handles disagreement and recovers correct minority answers. Its applicability to open-ended tasks, such as creative writing and multi-turn dialogue, remains to be evaluated. In these settings, disagreements may reflect complementary perspectives rather than factual errors. Extending \ourmethod to such scenarios would involve adapting relation assessment and aggregation criteria to task-specific goals, offering a natural direction for broader multi-agent collaboration.

\textbf{Potential Risks.} Improved multi-agent coordination may facilitate harmful automation or amplify shared errors and misleading outputs. Real-world deployment should therefore include permission controls and human oversight, particularly for consequential decisions.

\bibliography{ref}

\end{document}